\documentclass[11pt]{article}

\usepackage[final]{acl}

\usepackage{times}
\usepackage{latexsym}

\usepackage[T1]{fontenc}

\usepackage[utf8]{inputenc}

\usepackage{microtype}

\usepackage{inconsolata}

\usepackage{graphicx}

\usepackage{longtable}
\usepackage{booktabs}
\usepackage{siunitx} 
\usepackage{comment}

\title{Neural Machine Translation for Coptic-French: Strategies for Low-Resource Ancient Languages}

\author{Nasma Chaoui \and Richard Khoury \\
Department of Computer Science and Software Engineering \\
Université Laval \\
Québec City, Canada \\
\texttt{nasma.chaoui.1@ulaval.ca, richard.khoury@ift.ulaval.ca} \\
}

\begin{document}
\maketitle
\begin{abstract}
This paper presents the first systematic study of strategies for translating Coptic into French. 
Our comprehensive pipeline systematically evaluates: pivot versus direct translation, the impact of pre-training, the benefits of multi-version fine-tuning, and model robustness to noise. Utilizing aligned biblical corpora, we demonstrate that fine-tuning with a stylistically-varied and noise-aware training corpus significantly enhances translation quality. Our findings provide crucial practical insights for developing translation tools for historical languages in general.
\end{abstract}

\section{Introduction}
Neural Machine Translations (NMT), with its steady growth, has profoundly impacted the field of computational linguistics. Transformer models and large-scale multilingual systems have enabled translation coverage across hundreds of language pairs. However, many historical languages — especially ancient ones with scarce data — have yet to benefit from these advancements. Coptic is one such language. As the last stage of the Egyptian language family, understanding and translating it is crucial for advancing fields such as Egyptology and Early-Christian studies.

As a result, the language is studied in universities around the world; according to one listing\footnote{Listing assembled by \url{https://www.cmcl.it/~iacs/course.htm}.}, over 50 universities in 13 countries offer Coptic courses. But despite this international interest, there are currently few NMT tools for Coptic, and only to translate to English or Arabic. 

This paper aims to fill that gap by investigating a wide range of NMT strategies to translate Coptic into French. In the process, we will focus on four important questions: (1) should we translate directly to our target language or use English as a pivot language; (2) how should we pick a language model to fine-tune for this task; (3) should we provide the model with a diverse set of possible translations for a passage; and (4) how can we make the model robust to noise in the original manuscript. 
While our focus is solely on Coptic$\rightarrow$French NMT, these questions are of general interest, and our results will help guide the development of NMT tools between any ancient-modern language pair.

The remainder of this paper is organized as follows: Section \ref{sec:background} reviews relevant work on natural language processing Coptic. Section \ref{sec:metho} details our methodology we plan to use to explore our four questions. Section \ref{sec:expsetup} presents our experimental setup: our dataset, NMT models and metrics, and Section \ref{sec:resultsanalysis} presents the results we obtain on the experiments about each of the four questions. Finally, Section \ref{sec:conclusion} has some concluding remarks. Our dataset and code are available on our GitHub account\footnote{\url{https://github.com/chaouin/coptic-french-nmt}}, 
while the best Coptic$\rightarrow$French NMT models we train are distributed on HuggingFace\footnote{\url{https://huggingface.co/chaouin/coptic-french-translation-helsinki}}\footnote{\url{https://huggingface.co/chaouin/coptic-french-translation-hiero}}.

\section{Background}
\label{sec:background}

Coptic occupies a unique position as the final stage of the Egyptian language family, bridging thousands of years from Old Egyptian through Demotic. It holds critical historical and cultural significance for Egyptology, biblical and early monasticism studies, as many Coptic manuscripts illuminate these fields. 

In recent years, substantial progress has been made in building digital resources for Coptic. The Coptic SCRIPTORIUM project \cite{schroeder2016raiders} stands out as a foundational effort, providing annotated corpora with multiple linguistic layers, including part-of-speech tagging, lemmatization, morphological analysis, and syntactic parsing \cite{zeldes2016nlp}. 
Similarly, tools such as the BabyLemmatizer have extended lemmatization and POS tagging capabilities across Coptic, Demotic, and Earlier Egyptian, demonstrating the shared linguistic complexities within the Egyptian language family \cite{sahala-lincke-2025-neural}. 

Recent work has also explored NMT within the Egyptian language family, particularly focusing on hieroglyphic texts. Notably, \cite{cao2024deep} developed a Hieroglyphic Transformer by adapting the multilingual M2M-100 model to translate ancient Egyptian hieroglyphs into both German and English. Their approach involved constructing a carefully curated dataset from the \textit{Thesaurus Linguae Aegyptiae}, which includes hieroglyphic inscriptions paired with transliterations, POS tags, and translations. They relied on transliteration (romanization) to render hieroglyphic sequences compatible with modern NLP model vocabularies, and demonstrated that transfer learning from existing multilingual architectures could achieve promising results even on low-resource ancient scripts. 

The authors of \cite{saeed2024nile} introduced the largest parallel corpus to date for Coptic, comprising approximately 24,000 Coptic-English sentence pairs and 8,000 Coptic-Arabic pairs, and used it to train the first NMT system from Coptic to Arabic and English. They used both basic Seq2Seq models and fine-tuned multilingual models such as mT5 and M2M-100. 
Their work emphasized the necessity of romanization due to Coptic characters being absent from standard vocabularies. 
Similarly, \cite{enis2024ancient} implemented Coptic$\rightarrow$English NMT by fine-tuning 
multilingual models on normalized and romanized texts from the Coptic SCRIPTORIUM. They employed backtranslation and explored multiple data representations, demonstrating the critical impact of morphological normalization and tokenization for handling Coptic’s agglutinative structure.

\section{Methodology}
\label{sec:metho}

While implementing our Coptic$\rightarrow$French NMT model, we seek to study four fundamental design questions that are common to most if not all ancient-modern language translation projects. Thus, our methodology is divided into four stages, each building on the findings of the previous.

\subsection{Pertinence of Fine-Tuning}
\label{metho.ex1}
To begin, we seek to answer a fundamental question: is it even worthwhile to fine-tune a NMT model for Coptic$\rightarrow$French translation? After all, while no such model exists, there are alternatives. The first one would be to use a more general multilingual NMT model that can work out of the box without any fine-tuning. Another alternative could use a NMT model that can translate the ancient language into another pivot language, and prompt it to answer in our target modern language. In our case, this will mean using a readily-available Coptic$\rightarrow$English NMT model and prompting it to answer in French. Finally, since NMT models trained to translate between pairs of modern languages abound today, we could again start from an NMT model that can translate the ancient language into a pivot language, and use a second NMT model to translate that content into our target modern language. In our case, we could daisy-chain a Coptic$\rightarrow$English and an English$\rightarrow$French NMT model to accomplish this. Consequently, the first stage of our research consists in comparing the translations produced by a fine-tuned Coptic$\rightarrow$French NMT model and these three alternatives, to see the differences between these four options.

\subsection{Pre-Trained Model Choice}
\label{metho.ex2}
While there are not a lot of ancient$\rightarrow$modern NMT models, there are a handful of them available, and therefore a question arises: how to pick which one to fine-tune? There are technical and linguistic considerations in this decision. From a technical standpoint, the different models available may have different numbers of parameters and be trained using different quantities of data (but given the scarcity of ancient-language data, they will typically not have different data sources). And linguistically, when no model exists for our ancient and modern language pair, three options are available. The first is to take a model trained to translate the ancient language to a different modern language and fine-tune it to our target language, or in our case fine-tune a Coptic-English model to translate to French. The second option is to take a model trained to translate into the modern language and fine-tune it to the ancient language. In our case this would mean taking any of the available multilingual NMT models trained for French and fine-tune it to translate from Coptic. Finally, a NMT translation model may exist for a different ancient language in the same family, giving us a third option. As mentioned previously, Coptic is the final stage of the Egyptian language, so a NMT model trained for an earlier stage of the language could be fine-tuned for Coptic.

\subsection{Fine-Tuning Dataset Choice}
\label{metho.ex3}
If we want to fine-tune an existing model, we need a corpus of parallel text in the ancient and modern languages. However, for a variety of reasons, an ancient-language text may be translated to a modern language in several different ways. We thus conduct an experiment to see whether or not fine-tuning can benefit from this stylistic diversity. Our hypothesis is that having multiple different translations of a same original text will make it possible to build a larger training corpus from the scarce ancient language resource, while also allowing the NMT model to better learn concepts of syntactic variation and word synonymy. 

\subsection{Training Robustness to Noise}
\label{metho.ex4}
Ancient-language texts often suffer from corruption due to the passage of time. In our final experiment, we consider the benefit of training an NMT model using noisy data as a way of building in robustness to noise. We consider two sources of noise. The first are lacunae that result from missing or damaged fragments of the text \cite{levine-etal-2024-lacuna}. We can introduce this problem in our training data by randomly replacing characters in the text with a ``missing character'' symbol. The second source of noise in our study is letter substitutions. This problem results from two different sources: on the one hand certain characters in a language can look similar and hand-writing can be ambiguous, and on the other optical character recognition (OCR) systems used to digitize manuscripts are prone to misrecognize characters, with one study finding that character recognition accuracy in ancient text OCR could be as low as 62\% \cite{tzogka2021ocr}. We can simulate this problem by creating a list of similar-looking characters, and randomly replacing a character by a similar one from the list. A final source of error is spelling variations and mistakes, which we simulate by randomly switching the order of two adjacent characters. By controlling the rate of three three types of random replacements, we can further experiment with the impact of different proportions of noise in the data. 

\section{Experimental Setup}
\label{sec:expsetup}

\subsection{Dataset and Preprocessing}

To construct our parallel Coptic–French corpus, we first extracted Coptic texts from the \textit{Coptic SCRIPTORIUM} project \cite{schroeder2016raiders}, which provides richly annotated PAULA XML files. We focused on the Sahidic New and Old Testament Bibles, the most complete and standardized Coptic corpus available. We automatically parsed the XML files to extract verse-level Coptic content along with their verse ID (the biblical book, chapter and verse number).

Since most pre-trained NMT models are not compatible with the original Coptic Unicode script, we romanized the Coptic data using the \texttt{uroman} tool\footnote{\url{https://github.com/isi-nlp/uroman}}, an idea proposed by \cite{amrhein-sennrich-2020-romanization} and used for Coptic by \cite{enis2024ancient}. This ensures compatibility with the tokenizers and vocabularies of existing Transformer models.

For the French text, we collected three public-domain Bible translations: the Louis Segond, Crampon, and Darby versions. These were chosen for their stylistic diversity, modern relevance, and free availability. Alignment with the Coptic text was performed automatically using verse ID.

Next, we performed a cleaning step to ensure a high-quality parallel corpus. Any verse pair missing either the Coptic or French translation was eliminated. We also removed structurally mismatched pairs by filtering out entries where the Coptic segment consisted only of ellipses (e.g., "[...]"), indicating incomplete or missing verses. Additionally, we discarded verses whose French translation was missing or blank, and stripped leading inline annotations (e.g., (1.2)) from the French verses to preserve only the canonical content. All removed entries were logged separately to ensure reproducibility. 
After cleaning, we retained 23,561 aligned verses for both the Segond and Darby Bibles, and 23,718 verses for the Crampon version, for a total of 70,840 aligned pairs from 63 biblical books.

We split this dataset into a training and testing corpus. We opted for a book-level split, keeping the books of \textit{1 Corinthians}, \textit{Mark}, \textit{Galatians} and \textit{Hebrews} as test data and the other 59 books as training data. This represents a total of 1,460 Coptic test verses, each with 3 French translations. This setup prevents data contamination, where similar verses from a same book could be found in the training and testing corpora. It also simulates a real-world use case, of a new never-before-seen Coptic document being discovered.

\subsection{Pre-trained NMT Models}
\label{sec:models}
We selected four pre-trained NMT models for our study. They were the most related ones to our challenge of Coptic$\rightarrow$French translation. 
\begin{itemize}
    
    \item \textbf{Enis}: The first-ever Coptic$\rightarrow$English NMT model, by \cite{enis2024ancient}\footnote{\url{https://huggingface.co/megalaa/coptic-english-translator}}.

    \item \textbf{Helsinki}: A multilingual model from the University of Helsinki \cite{tiedemann2023democratizing} trained for over 100 languages using a massive parallel Bible corpus, including Coptic (using a New Testament text different from the SCRIPTORIUM version) and French\footnote{\url{https://huggingface.co/Helsinki-NLP/opus-mt-tc-bible-big-mul-mul}}.

    \textbf{T5}: A general-purpose encoder–decoder model trained by Google on the C4 corpus \cite{raffel2020exploring}, with exposure to French. It does not include Coptic. \footnote{\url{https://huggingface.co/t5-base}}

    \item \textbf{Hiero}: A NMT model fine-tuned by  \cite{cao2024deep} for Hieroglyphic$\rightarrow$English and Hieroglyphic$\rightarrow$German translation\footnote{\url{https://huggingface.co/mattiadc/hiero-transformer}}.
    
\end{itemize}

Unless specified differently, all models were fine-tuned for 15 epochs on our training corpus.

\subsection{Metrics}
\label{sec:metrics}

As our dataset clearly demonstrates, a single Coptic statement can have several different but equivalent French translations. 
Thus, a good model is one that correctly translates the intended meaning of the original statement, even if the wording differs from the gold-standard reference.
To evaluate translation quality, we focus on metrics that measure semantic similarity rather than lexical overlap.

\begin{itemize}
    \item \textbf{BERTScore} \cite{bert-score} computes token-level similarity using contextual embeddings from a pre-trained BERT model. Scores range from 0 to 1, where higher values indicate greater semantic similarity between the candidate and reference sentences. It is well-suited to evaluate semantic equivalence, particularly when multiple valid versions of a statement exist.
    
    \item \textbf{BLEURT} \cite{sellam2020bleurt} is a learned evaluation metric that fine-tunes BERT-based models to predict human judgments of text quality. It combines semantic similarity, fluency, and grammaticality into a single score, and has been widely used to evaluate machine translation outputs \cite{pu2021learning}. BLEURT scores are unbounded but typically center around [0, 1], with higher values indicating better quality.
    
    \item \textbf{COMET} \cite{rei-etal-2020-comet}
    is a learned evaluation metric specifically designed for machine translation. It trains a multilingual BERT model to predict human judgment of translation quality. 
    COMET scores are in the range of 0 to 1, with higher scores correlating with higher human appreciation of the translation. 
    
    \item \textbf{METEOR} \cite{banarjee2005} is a machine translation metric based on unigram matching, but that takes into account synonymy, stemming, and paraphrases, and that has been adapted to reward semantically-valid phrasings rather than exact lexical matches. METEOR scores are normalized between 0 and 1, with 1 indicating a perfect match.
\end{itemize}

By considering these metrics together, we can obtain a more holistic evaluation of translation quality, one that accounts for meaning preservation and text quality and is tolerant of semantically-equivalent lexical changes.

\section{Results and Analysis}
\label{sec:resultsanalysis}

\subsection{Baseline}
\label{results.ex0}
We begin by establishing a set of baselines using the models of section \ref{sec:models}. To do this, each model is evaluated by translating our test corpus books in the most relevant language pair they are trained for: Coptic$\rightarrow$English for Enis, Coptic$\rightarrow$French for Helsinki, and English$\rightarrow$French for T5. As no Biblical texts exist in hieroglyphic form to evaluate the Hiero model, we tested it using its own testing corpus sentences, which is composed of 50 Hieroglyph$\rightarrow$English pairs and 50 Hieroglyph$\rightarrow$German pairs.

\begin{table}[ht]
  \centering
  \resizebox{\linewidth}{!}{%
  \begin{tabular}{lcccc}
    \hline
    \textbf{Model} & \textbf{BERTScore} & \textbf{BLEURT} & \textbf{COMET} & \textbf{METEOR} \\
    \hline
    Enis & 0.632 & 0.344 & 0.328 & 0.119 \\
    Helsinki & 0.595 & 0.136 & 0.248 & 0.082 \\
    T5 & 0.809 & 0.501 & 0.673 & 0.429 \\
    Hiero & 0.833 & 0.555 & 0.664 & 0.496 \\
    \hline
  \end{tabular}
  }
  \caption{Baseline pre-trained model performances.}
  \label{tab:baselines}
\end{table}

\subsection{Pertinence of Fine-Tuning}
\label{results.ex1}

Our first experiment compares the four different strategies to achieve Coptic$\rightarrow$French translation described in Section \ref{metho.ex1}. First is the strategy of fine-tuning an existing Coptic translation model, the Enis model, to a French target. The second strategy is to use a multilingual model that includes both Coptic and French out of the box, namely the Helsinki model. The third strategy is to prompt the Enis model to translate to French without fine-tuning, in a zero-shot manner. Finally, we will use English as a pivot language, by using Enis to translate from Coptic to English then translating that output to French using T5.

\begin{table}[ht]
  \centering
  \resizebox{\linewidth}{!}{%
  \begin{tabular}{lcccc}
    \hline
    \textbf{Approach} & \textbf{BERTScore} & \textbf{BLEURT} & \textbf{COMET} & \textbf{METEOR} \\
    \hline
    Fine-tuning & \textbf{0.820} & \textbf{0.402} & \textbf{0.613} & \textbf{0.467} \\
    Multilingual & 0.595 & 0.136 & 0.248 & 0.082 \\
    Prompting & 0.606 & 0.149 & 0.238 & 0.041 \\
    Pivot language & 0.620 & 0.115 & 0.274 & 0.100 \\
    \hline
  \end{tabular}
  }
  \caption{Comparison of translation strategies.}
  \label{tab:exp1}
\end{table}

The results in Table \ref{tab:exp1} clearly demonstrates that the benefits of fine-tuning justify the extra labor, compared to the other three methods.  
The Enis model generates better translations in French after being fine-tuned than it did in English in our baseline results in Table \ref{tab:baselines}, thanks to the extra training. Prompting Enis to translate to French without fine-tuning gives the worst results of the three Enis tests, due to mismatched decoding expectations.  
The model fine-tuned to a single pair of languages also outperforms the multilingual model, a result coherent with the literature, which finds that multiligual models perform suboptimally for low-resource languages (such as Coptic) \cite{joshi-etal-2025-adapting}. Finally, it is interesting to see that the Enis-T5 pair achieves worse results on all metrics than either Enis or T5 did in the baseline test of Table \ref{tab:baselines}. This indicates that translation errors in each step are compounded by the sequence of translations, reducing overall accuracy. This matches the observations of \cite{paul2013choose}.

\subsection{Pre-Trained Model Choice}
\label{results.ex2}

Having confirmed that fine-tuning a model for Coptic$\rightarrow$French is worth the effort, our second experiment explores the question of Section \ref{metho.ex2}, namely which model should be fine-tuned. For this experiment, we fine-tune each model on our entire training corpus. This enables Enis to learn to translate to French, T5 to translate from Coptic, and Hiero to translate from Coptic to French. While Helsinki was already trained on Coptic and French among its 100 languages, it was also fine-tuned to become specialized to those two languages.

\begin{table}[ht]
  \centering
  \resizebox{\linewidth}{!}{%
  \begin{tabular}{lcccc}
    \hline
    \textbf{Model} & \textbf{BERTScore} & \textbf{BLEURT} & \textbf{COMET} & \textbf{METEOR} \\
    \hline
    Enis & 0.820 & 0.402 & 0.613 & 0.467 \\
    Helsinki & \textbf{0.850} & \textbf{0.564} & \textbf{0.731} & \textbf{0.563} \\
    T5 & 0.763 & 0.240 & 0.474 & 0.325 \\
    Hiero & 0.832 & 0.498 & 0.681 & 0.501 \\
    \hline
  \end{tabular}
  }
  \caption{Comparison of fine-tuned models.}
  \label{tab:exp2}
\end{table}

The results translating our test corpus, presented in Table \ref{tab:exp2}, show that Helsinki outperforms the other models on all four metrics.
This performance can be attributed to two key advantages. First, its pre-training corpus includes both the source and target languages, unlike the other models that lacked one or both of the languages and had to learn them from scratch.
Second, it's the model whose pre-training corpus is most aligned with our task, being also a Bible corpus. The Helsinki model is also relatively large, with 248M parameters, which may contribute to its strong performance. However, we cannot exclude the possibility that it is becomming over-fitted to translating biblical texts; since we do not have non-biblical Coptic-French translations to test on, we cannot check that possibility. 

The next best model on all metrics is Hiero. Interestingly, this one has the opposite properties from Helsinki: it is trained on neither Coptic, French, nor biblical text. It is however trained for Hieroglyphs, a language related to Coptic. This seems to indicate that knowledge learned on a language transfers well to related languages. A similar observation was made by \cite{babych-etal-2007-translating} in the case of NMT using pivot languages. The Enis model is a close third place. While that model benefits from having been pre-trained on Coptic, it is smaller than the Hiero model, with 77.1M against 484M parameters for Hiero, and thus doesn't learn the new language as well. Finally, the T5 model, with 223M parameters, which does not have any Coptic nor related language in its pre-training data, achieves the worst fine-tuning results. 

\subsection{Fine-Tuning Dataset Choice}
\label{results.ex3}

To test our hypothesis of Section \ref{metho.ex3}, we fine-tune the two best models of Section \ref{results.ex2}, Helsinki and Hiero, on each of our three French translations individually. For this experiment, since each single-source model trains on roughly one-third of the complete training dataset, we train them for 45 epochs instead of 15 for the model that trains on the entire dataset. We tested each model on each translation of the test corpus separately, to observe the impact each different training has.

\begin{table}[ht]
  \centering
  \resizebox{\linewidth}{!}{%
  \begin{tabular}{llcccc}
    \hline
    \textbf{Test set} & \textbf{Training set} & \textbf{BERTScore} & \textbf{BLEURT} & \textbf{COMET} & \textbf{METEOR} \\
    \hline
    Crampon & All datasets     & 0.836 & \textbf{0.549} & \textbf{0.723} & 0.513 \\
    Crampon & Crampon only     & \textbf{0.849} & 0.537 & 0.715 & \textbf{0.583} \\
    Crampon & Darby only       & 0.821 & 0.494 & 0.690 & 0.460 \\
    Crampon & Segond only      & 0.825 & 0.519 & 0.703 & 0.472 \\
    \hline
    Darby   & All datasets     & 0.858 & \textbf{0.576} & \textbf{0.737} & 0.587 \\
    Darby   & Crampon only     & 0.825 & 0.511 & 0.696 & 0.493 \\
    Darby   & Darby only       & \textbf{0.868} & 0.558 & 0.729 & \textbf{0.615} \\
    Darby   & Segond only      & 0.846 & 0.535 & 0.710 & 0.546 \\
    \hline
    Segond  & All datasets     & 0.856 & \textbf{0.569} & \textbf{0.732} & 0.587 \\
    Segond  & Crampon only     & 0.828 & 0.517 & 0.699 & 0.514 \\
    Segond  & Darby only       & 0.844 & 0.515 & 0.701 & 0.550 \\
    Segond  & Segond only      & \textbf{0.865} & 0.564 & 0.727 & \textbf{0.611} \\
    \hline
  \end{tabular}
  }
  \caption{Comparison of dataset choice on Helsinki.}
  \label{tab:exp3_detail_opus}
\end{table}

\begin{table}[ht]
  \centering
  \resizebox{\linewidth}{!}{%
  \begin{tabular}{llcccc}
    \hline
    \textbf{Test set} & \textbf{Training set} & \textbf{BERTScore} & \textbf{BLEURT} & \textbf{COMET} & \textbf{METEOR} \\
    \hline
    Crampon & All datasets & 0.815 & \textbf{0.480} & \textbf{0.672} & 0.445 \\
    Crampon & Crampon only & \textbf{0.825} & 0.458 & 0.655 & \textbf{0.493} \\
    Crampon & Darby only   & 0.805 & 0.433 & 0.642 & 0.410 \\
    Crampon & Segond only  & 0.806 & 0.448 & 0.646 & 0.413 \\
    \hline
    Darby   & All datasets & 0.844 & \textbf{0.517} & \textbf{0.692} & 0.535 \\
    Darby   & Crampon only & 0.812 & 0.450 & 0.648 & 0.444 \\
    Darby   & Darby only   & \textbf{0.847} & 0.493 & 0.678 & \textbf{0.547} \\
    Darby   & Segond only  & 0.830 & 0.472 & 0.660 & 0.489 \\
    \hline
    Segond  & All datasets & 0.837 & \textbf{0.495} & \textbf{0.680} & 0.519 \\
    Segond  & Crampon only & 0.810 & 0.442 & 0.642 & 0.443 \\
    Segond  & Darby only   & 0.826 & 0.451 & 0.648 & 0.487 \\
    Segond  & Segond only  & \textbf{0.839} & 0.479 & 0.663 & \textbf{0.521} \\
    \hline
  \end{tabular}
  }
  \caption{Comparison of dataset choice on Hiero.}
  \label{tab:exp3_detail_hiero}
\end{table}

The results, given in Tables \ref{tab:exp3_detail_opus} and \ref{tab:exp3_detail_hiero}, show that the models trained on a specific dataset translation perform best when tested on that same translation in two of our four metrics, BERTScore and METEOR. These are the two metrics that value semantic similarity to the correct translation. By training and testing on the same translation, these models learn the correct choice of wording and translation decisions, and thus can generate a translation that is very similar to the expected one. By contrast, when training on one translation and testing on another, these systems all yield much weaker BERTScore and METEOR performances, showing they are generating translations worded differently from the expected result. 

On the other hand, the model trained on all three dataset translations always achieves the top BLEURT and COMET scores, no matter which dataset translation it is tested on. These two metrics are tuned to replicate human appreciation of the generated text's quality. By training on a variety of possible translations for each Coptic sentence, that model has learned to generate better translations overall. In addition, this model always achieves the second-best BERTScore and METEOR scores, behind the model trained on the same test corpus translation but ahead of the two trained on other translations. So while the wording generated by this model is not optimal compared to the correct translation, it is always closer than any other model not specialized to generate that specific translation. Overall, this shows that stylistic diversity in the training data improves generation quality at inference time, especially when multiple valid translations are acceptable.

\subsection{Training Robustness to Noise}
\label{results.ex4}
\begin{figure}[h]
  \centering
  \includegraphics[width=0.8\linewidth]{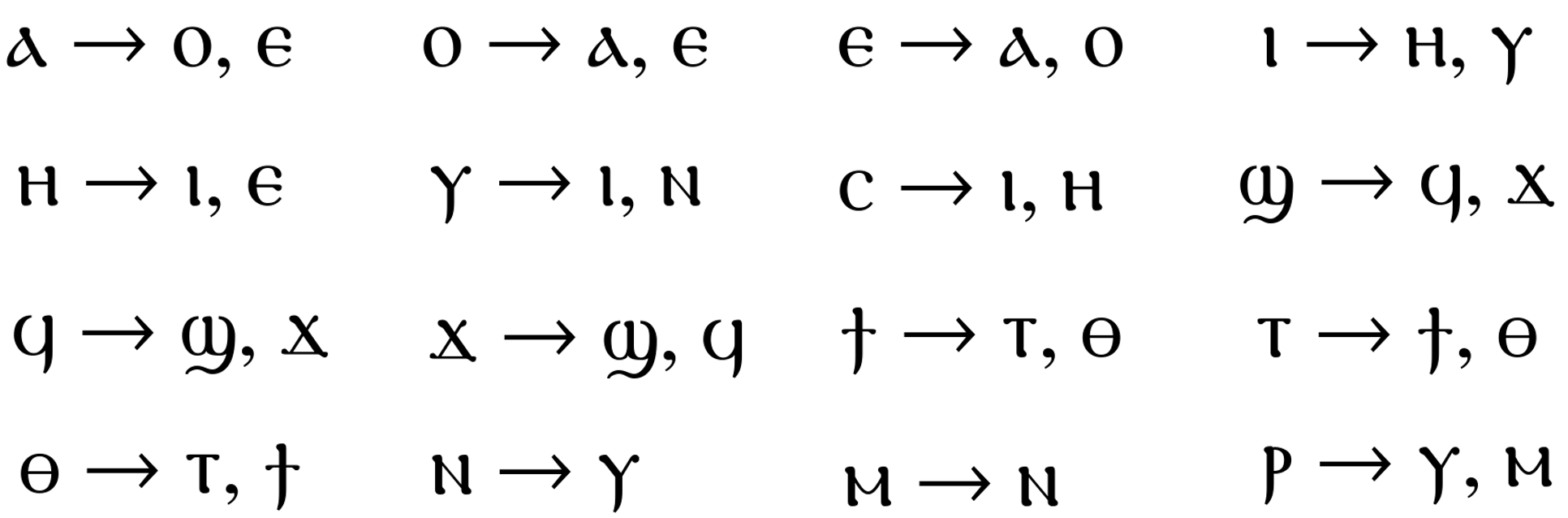}
  \caption{Visual confusion of Coptic characters.}
  \label{fig:coptic-confusion-map}
\end{figure}

As explained in Section \ref{metho.ex4}, we simulate real-world degradation of the manuscripts by injecting noise in our training and testing data. To simulate textual corruption, each character in a Coptic verse is subjected to three independent noise processes: a 2\% probability of deletion (simulating lacunae), a 2\% probability of switch with an adjacent character (simulating spelling errors), and a 10\% probability of replacement by a visually similar character (simulating OCR errors), using the visual confusion mapping of Figure \ref{fig:coptic-confusion-map}. 
We do not alter the French translations of the verses. We experiment with different levels of simulated degradation by creating 5 versions of our corpora, setting the probability that a verse is corrupted to 0\%, 10\%, 30\%, 50\%, and 100\%. 

\begin{table}[ht]
  \centering
  \resizebox{\linewidth}{!}{%
  \begin{tabular}{llcccc}
    \hline
    \textbf{Test noise} & \textbf{Train noise} & \textbf{BERTScore} & \textbf{BLEURT} & \textbf{COMET} & \textbf{METEOR} \\
    \hline
    0\% & 0\%   & 0.850 & 0.564 & 0.731 & 0.563 \\
    0\% & 10\%  & \textbf{0.852} & \textbf{0.568} & \textbf{0.733} & \textbf{0.564} \\
    0\% & 30\%  & 0.849 & 0.560 & 0.727 & 0.560 \\
    0\% & 50\%  & 0.848 & 0.553 & 0.723 & 0.559 \\
    0\% & 100\% & 0.846 & 0.547 & 0.719 & 0.546 \\
    \hline
    10\% & 0\%   & 0.844 & 0.545 & 0.717 & 0.549 \\
    10\% & 10\%  & \textbf{0.849} & \textbf{0.557} & \textbf{0.725} & \textbf{0.557} \\
    10\% & 30\%  & 0.847 & 0.553 & 0.722 & 0.555 \\
    10\% & 50\%  & 0.846 & 0.545 & 0.718 & 0.554 \\
    10\% & 100\% & 0.844 & 0.544 & 0.717 & 0.543 \\
    \hline
    30\% & 0\%   & 0.829 & 0.492 & 0.679 & 0.511 \\
    30\% & 10\%  & 0.842 & 0.534 & 0.708 & 0.538 \\
    30\% & 30\%  & 0.842 & \textbf{0.539} & 0.711 & 0.545 \\
    30\% & 50\%  & \textbf{0.843} & 0.536 & \textbf{0.712} & \textbf{0.546} \\
    30\% & 100\% & 0.841 & 0.533 & 0.709 & 0.536 \\
    \hline
    50\% & 0\%   & 0.817 & 0.457 & 0.650 & 0.478 \\
    50\% & 10\%  & 0.837 & 0.514 & 0.695 & 0.522 \\
    50\% & 30\%  & \textbf{0.838} & 0.522 & 0.701 & \textbf{0.533} \\
    50\% & 50\%  & \textbf{0.838} & 0.519 & 0.698 & \textbf{0.533} \\
    50\% & 100\% & \textbf{0.838} & \textbf{0.524} & \textbf{0.702} & 0.529 \\
    \hline
    100\% & 0\%   & 0.788 & 0.366 & 0.579 & 0.411 \\
    100\% & 10\%  & 0.822 & 0.471 & 0.665 & 0.488 \\
    100\% & 30\%  & 0.826 & 0.491 & 0.675 & 0.502 \\
    100\% & 50\%  & 0.829 & 0.496 & 0.683 & 0.509 \\
    100\% & 100\% & \textbf{0.831} & \textbf{0.505} & \textbf{0.688} & \textbf{0.513} \\
    \hline
  \end{tabular}
  }
  \caption{Comparison of robustness to noise for Helsinki.}
  \label{tab:exp4_detail_opus}
\end{table}

\begin{table}[ht]
  \centering
  \resizebox{\linewidth}{!}{%
  \begin{tabular}{llcccc}
    \hline
    \textbf{Test noise} & \textbf{Train noise} & \textbf{BERTScore} & \textbf{BLEURT} & \textbf{COMET} & \textbf{METEOR} \\
    \hline
    0\% & 0\%   & \textbf{0.832} & 0.497 & \textbf{0.682} & \textbf{0.500} \\
    0\% & 10\%  & \textbf{0.832} & 0.496 & 0.679 & 0.496 \\
    0\% & 30\%  & 0.831 & \textbf{0.498} & 0.679 & 0.494 \\
    0\% & 50\%  & 0.830 & 0.492 & 0.674 & 0.492 \\
    0\% & 100\% & 0.827 & 0.482 & 0.668 & 0.482 \\
    \hline
    10\% & 0\%   & 0.825 & 0.474 & 0.665 & 0.485 \\
    10\% & 10\%  & 0.828 & 0.484 & 0.670 & 0.488 \\
    10\% & 30\%  & \textbf{0.829} & \textbf{0.489} & \textbf{0.674} & \textbf{0.489} \\
    10\% & 50\%  & \textbf{0.829} & 0.485 & 0.671 & 0.488 \\
    10\% & 100\% & 0.826 & 0.478 & 0.665 & 0.480 \\
    \hline
    30\% & 0\%   & 0.809 & 0.412 & 0.619 & 0.445 \\
    30\% & 10\%  & 0.822 & 0.452 & 0.649 & 0.470 \\
    30\% & 30\%  & \textbf{0.825} & 0.465 & \textbf{0.658} & \textbf{0.476} \\
    30\% & 50\%  & 0.824 & \textbf{0.466} & 0.656 & \textbf{0.476} \\
    30\% & 100\% & 0.824 & \textbf{0.466} & 0.657 & 0.473 \\
    \hline
    50\% & 0\%   & 0.795 & 0.364 & 0.582 & 0.416 \\
    50\% & 10\%  & 0.815 & 0.426 & 0.630 & 0.453 \\
    50\% & 30\%  & 0.819 & 0.447 & 0.643 & 0.463 \\
    50\% & 50\%  & \textbf{0.821} & 0.452 & 0.647 & \textbf{0.466} \\
    50\% & 100\% & \textbf{0.821} & \textbf{0.456} & \textbf{0.650} & \textbf{0.466} \\
    \hline
    100\% & 0\%   & 0.762 & 0.232 & 0.486 & 0.335 \\
    100\% & 10\%  & 0.800 & 0.366 & 0.592 & 0.411 \\
    100\% & 30\%  & 0.810 & 0.405 & 0.616 & 0.436 \\
    100\% & 50\%  & 0.813 & 0.419 & 0.627 & 0.445 \\
    100\% & 100\% & \textbf{0.815} & \textbf{0.426} & \textbf{0.629} & \textbf{0.451} \\
    \hline
  \end{tabular}
  }
  \caption{Comparison of robustness to noise for Hiero.}
  \label{tab:exp4_detail_hiero}
\end{table}

The results in Tables \ref{tab:exp4_detail_opus} and \ref{tab:exp4_detail_hiero} show several clear trends. First, both models trained with any level of noise perform better on clean test data, and their scores drop steadily as noise increases. This shows that texts with errors or gaps are naturally harder to translate than clean ones.
Likewise, both the Helsinki and Hiero models trained with low noise do better on low-noise data, while those trained with more noise perform better on noisy data. For Helsinki, the model trained with 10\% noise leads early on, then the 50\% model takes over in mid-noise settings, and the 100\% model does best at the end. For Hiero, the 0\% model performs best early, the 30\% model leads in the middle, and the 100\% model dominates in the last tests.

What’s really happening is that models trained with less noise do better on clean data, but they’re also more fragile: their scores drop sharply as noise rises. Table \ref{tab:exp4drop} shows, for each model and metric, the relative score drop between 0\% and 100\% test noise. Models trained with 0\% and 10\% noise lose around 20\% on average—Hiero's BLEURT score even drops by 53\%. In contrast, models trained with 50\% or 100\% noise drop less than 15\%, and just 6\% on average. This confirms that controlled noise injection creates NMT models more resilient to real-world degradation, such as OCR errors or fragmentary manuscripts, at the cost of lower performances on clean manuscripts. While the optimal choice thus depends on the characteristics of the manuscript being translated, a good general-case compromise value seems to be at 50\% training noise.

\begin{table}[ht]
  \centering
  \resizebox{\linewidth}{!}{%
  \begin{tabular}{llcccc}
    \hline
    \textbf{Model} & \textbf{Training noise} & \textbf{BERTScore} & \textbf{BLEURT} & \textbf{COMET} & \textbf{METEOR} \\
    \hline
    Helsinki & 0\% & 7.3\% & 35.1\% & 20.8\% & 27.0\% \\
    Helsinki & 10\% & 3.5\% & 17.1\% & 9.3\% & 13.5\% \\
    Helsinki & 30\% & 2.7\% & 12.3\% & 7.2\% & 10.4\% \\
    Helsinki & 50\% & 2.2\% & 10.3\% & 5.5\% & 8.9\% \\
    Helsinki & 100\% & 1.8\% & 7.7\% & 4.3\% & 6.0\% \\
    \hline
    Hiero & 0\% & 8.4\% & 53.3\% & 28.7\% & 33.0\% \\
    Hiero & 10\% & 3.8\% & 26.2\% & 12.8\% & 17.1\% \\
    Hiero & 30\% & 2.5\% & 18.7\% & 9.3\% & 11.7\% \\
    Hiero & 50\% & 2.0\% & 14.8\% & 7.0\% & 9.6\% \\
    Hiero & 100\% & 2\% & 11.6\% & 5.8\% & 6.4\% \\
    \hline
  \end{tabular}
  }
  \caption{Drop in performance from 0\% to 100\% noise.}
  \label{tab:exp4drop}
\end{table}

\section{Conclusion}
\label{sec:conclusion}

This paper presents the first systematic study of neural machine translation for the Coptic$\rightarrow$French language pair. Through four experiments, we outlined best practices for building a translation model for this new pair.
First, we showed that task-specific fine-tuning clearly improves translation quality compared to using pre-trained multilingual models or translating through a pivot language. Next, we found that fine-tuning a model already trained on the ancient language or a related one in the same family (such as Hieroglyphics) yields better results. Third, we examined the effect of having multiple translations of the same text, and found that including them all enhanced translation quality. Finally, we tested the impact of noise in the dataset and found that the best trade-off between translation quality and robustness was achieved when training with 50\% corrupted verses.
This led to the release of two strong NMT models for this task, one based on the multilingual model of \cite{tiedemann2023democratizing} and the other on the Hieroglyphic translator of \cite{cao2024deep}, both fine-tuned on three French versions of the training corpus with 50\% noise.
Together, our findings extend beyond Coptic$\rightarrow$French translation, providing practical guidance for building robust NMT systems in other low-resource or ancient language settings.

\section*{Acknowledgements}
This research was made possible by funding from the Natural Sciences and Engineering Research Council of Canada (NSERC). 

\bibliography{custom}

\appendix

\end{document}